Original Paper

# Aligning Large Language Models for Enhancing Psychiatric Interviews Through Symptom Delineation and Summarization: Pilot Study

Jae-hee So[1], BA; Joonhwan Chang[1], BA; Eunji Kim[2,3], MA; Junho Na[1], BA; JiYeon Choi[4,5], RN, PhD; Jy-yong Sohn[1*], PhD; Byung-Hoon Kim[2,3,5,6*], MD, PhD; Sang Hui Chu[4,5*], RN, PhD

[1]Department of Applied Statistics, Yonsei University, Seoul, Republic of Korea
[2]Department of Psychiatry, Yonsei University College of Medicine, Seoul, Republic of Korea
[3]Institute of Behavioral Sciences in Medicine, Yonsei University College of Medicine, Seoul, Republic of Korea
[4]Department of Nursing, Mo-Im Kim Nursing Research Institute, Yonsei University College of Nursing, Seoul, Republic of Korea
[5]Institute for Innovation in Digital Healthcare, Yonsei University, Seoul, Republic of Korea
[6]Department of Biomedical Systems Informatics, Yonsei University College of Medicine, Seoul, Republic of Korea
*these authors contributed equally

**Corresponding Author:**
Jy-yong Sohn, PhD
Department of Applied Statistics
Yonsei University
50 Yonsei-ro
Seodaemun-gu
Seoul, 03722
Republic of Korea
Phone: 82 2 2123 2472
Email: jysohn1108@gmail.com

## Abstract

**Background:** Recent advancements in large language models (LLMs) have accelerated their use across various domains. Psychiatric interviews, which are goal-oriented and structured, represent a significantly underexplored area where LLMs can provide substantial value. In this study, we explore the application of LLMs to enhance psychiatric interviews by analyzing counseling data from North Korean defectors who have experienced traumatic events and mental health issues.

**Objective:** This study aims to investigate whether LLMs can (1) delineate parts of the conversation that suggest psychiatric symptoms and identify those symptoms, and (2) summarize stressors and symptoms based on the interview dialogue transcript.

**Methods:** Given the interview transcripts, we align the LLMs to perform 3 tasks: (1) extracting stressors from the transcripts, (2) delineating symptoms and their indicative sections, and (3) summarizing the patients based on the extracted stressors and symptoms. These 3 tasks address the 2 objectives, where delineating symptoms is based on the output from the second task, and generating the summary of the interview incorporates the outputs from all 3 tasks. In this context, the transcript data were labeled by mental health experts for the training and evaluation of the LLMs.

**Results:** First, we present the performance of LLMs in estimating (1) the transcript sections related to psychiatric symptoms and (2) the names of the corresponding symptoms. In the zero-shot inference setting using the GPT-4 Turbo model, 73 out of 102 transcript segments demonstrated a recall mid-token distance d<20 for estimating the sections associated with the symptoms. For evaluating the names of the corresponding symptoms, the fine-tuning method demonstrates a performance advantage over the zero-shot inference setting of the GPT-4 Turbo model. On average, the fine-tuning method achieves an accuracy of 0.82, a precision of 0.83, a recall of 0.82, and an F1-score of 0.82. Second, the transcripts are used to generate summaries for each interviewee using LLMs. This generative task was evaluated using metrics such as Generative Evaluation (G-Eval) and Bidirectional Encoder Representations from Transformers Score (BERTScore). The summaries generated by the GPT-4 Turbo model, utilizing both symptom and stressor information, achieve high average G-Eval scores: coherence of 4.66, consistency of 4.73, fluency of 2.16, and relevance of 4.67. Furthermore, it is noted that the use of retrieval-augmented generation did not lead to a significant improvement in performance.

**Conclusions:** LLMs, using either (1) appropriate prompting techniques or (2) fine-tuning methods with data labeled by mental health experts, achieved an accuracy of over 0.8 for the symptom delineation task when measured across all segments in the transcript. Additionally, they attained a G-Eval score of over 4.6 for coherence in the summarization task. This research contributes to the emerging field of applying LLMs in psychiatric interviews and demonstrates their potential effectiveness in assisting mental health practitioners.

## Introduction

Globally, the demand for mental health services is substantial and continues to grow, underscoring the increasing need for support and resources to address mental health challenges. In 2010, the social cost of poor mental health worldwide was estimated at approximately US $2.5 trillion annually, with this cost projected to more than double by 2030 [1]. However, access to and engagement with mental health care services remain hindered by factors such as high costs and a shortage of mental health specialists [2]. In recent years, particularly after the COVID-19 pandemic, digital health care and artificial intelligence (AI) have gained traction as alternatives to overcome these limitations by enhancing the clinical efficiency of mental health care professionals [3]. Among the many potential applications of AI in improving the clinical workflow, most psychiatrists recognize that documenting medical records and synthesizing information will become key technologies in the near future [4].

Meanwhile, the rapid advancement of large language models (LLMs) [5-14] in the field of AI is transforming various industries. Although LLMs are typically pretrained on large corpora of unlabeled text using tasks such as next-token prediction [6] or masked language modeling [5], they exhibit the emergent ability to solve zero-shot tasks that they were not explicitly trained for [7,9]. Furthermore, fine-tuning these pretrained LLMs with a small set of labeled data, or aligning them at inference time using natural language prompting techniques, can enable LLMs to perform remarkably well on specific tasks [8]. Some widely known prompting techniques that enhance LLM performance include in-context learning [15], chain-of-thought reasoning [16], and other approaches [17-19]. These techniques assist LLMs by providing a small set of task-specific examples or guiding them through a structured reasoning process to solve the task.

In light of these advancements, numerous studies have explored the use of LLMs in medicine. Recent research consistently affirms the efficacy of LLMs in health care settings [20-27]. Alongside these findings, growing evidence suggests that LLMs can perform exceptionally well on clinical tasks beyond structured clinical question answering, such as clinical text summarization, when appropriate techniques are used to align the models [28,29].

Considering that psychiatric evaluation and intervention often involve intensive linguistic interviews between the patient and the psychiatrist, specific applications of LLMs in psychiatry are attracting increasing interest from researchers [30,31]. For example, a study by Galatzer-Levy et al [32] demonstrated that Medical-Pathways Language Model (Med-PaLM) 2 [33] could reasonably predict clinical scale scores based on clinical descriptions and interview dialogues. Another study by Luykx et al [34] evaluated ChatGPT's ability to answer clinical questions in psychiatry, showing that it could respond with high accuracy, completeness, and nuance. Additionally, clinical diagnosis matching for patients with psychiatric problems, based on the history of present illness using an electronic health record fine-tuned Bidirectional Encoder Representations from Transformers (BERT) model, demonstrated performance comparable to that of residents and semidesignated psychiatrists [35]. Although these studies provide empirical evidence that LLMs can be useful in clinical psychiatry, little research has been conducted on their application for summarizing medical records and synthesizing information—tasks that psychiatrists believe could significantly improve clinical workflow efficiency [4].

Aligned with these expectations, we explore the potential use of LLMs to enhance psychiatric interviews. Specifically, we define 2 research questions (RQs) that are closely related to improving clinical workflow in practice:

- RQ1: Can LLMs (1) identify which parts of a patient's utterances are related to psychiatric symptoms and (2) accurately name the corresponding symptoms?
- RQ2: Can LLMs effectively summarize stressors and symptoms from an interview between a patient with posttraumatic stress disorder (PTSD) and a trained interviewer?

If RQ1 can be answered, clinicians will be better able to identify important verbatim expressions from patients and assess the reliability of the LLM's output. Additionally, if RQ2 is answered, psychiatrists can more easily review patients' significant histories after interviews, saving time on clinical record documentation. To summarize and predict experiences and symptoms related to mental disorders based on counseling records, it is essential to utilize data in which these symptoms and experiences are explicitly evident. Consequently, because the experiences of mental disorders are most prominently manifested in cases of PTSD, our study focuses on experimenting with this population. To address these RQs, we utilize a curated set of interview transcripts from 10 North Korean defectors who have experienced significant stressors and trauma before, during, and after their displacement. These transcripts, labeled by mental health professionals, are used to explore the potential use of LLMs in enhancing psychiatric interviews.

Our main contributions are as follows:

- We evaluate LLMs on their ability to identify parts of the interview transcript that indicate psychiatric symptoms and to predict the types of symptoms. Our experimental results demonstrate that LLMs can effectively determine which portions of the dialogue convey psychiatric symptoms, as measured by the recall mid-token distance metric, a new metric we proposed based on clinical considerations.
- We evaluate LLMs on their ability to summarize the stressors and symptoms of interviewee patients. Our results indicate high performance in interview summarization—achieved through appropriate prompting and retrieval-augmented generation (RAG)—measured by the Generative Evaluation (G-Eval) [36] and Bidirectional Encoder Representations from Transformers Score (BERTScore) [37] metrics.

We expect our empirical results to provide initial guidance for researchers exploring techniques to adapt LLMs for clinical psychiatry applications. Figure 1 illustrates how our proposed method can facilitate synthesized information and documentation during the interview process.

**Figure 1.** Comparison between conventional and proposed methods for diagnosing the patients' mental disorders.

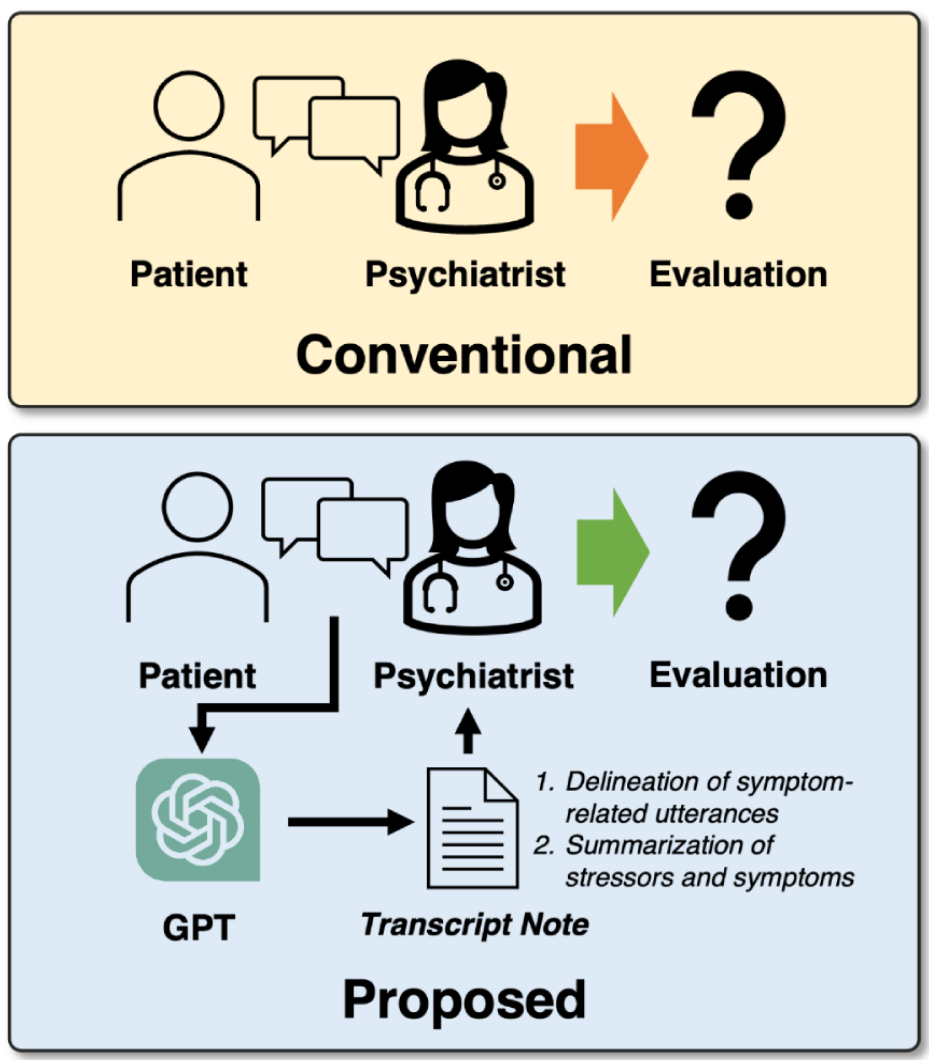

## Methods

### Data Set Acquisition

The study included 10 sets of interview transcripts derived from a qualitative investigation into the participants' traumatic experiences, symptoms, and the subsequent impact on their daily lives. In this study, we focused on North Korean defectors residing in South Korea who (1) were aged between 19 and 65 years and (2) reported experiencing 2 or more traumatic events along with posttraumatic stress symptoms, scoring 33 or higher on the Diagnostic and Statistical Manual of Mental Disorders, Fifth Edition (DSM-5)'s Posttraumatic Stress Disorder Checklist (PCL-5). The exclusion criteria were as follows: (1) a history of psychiatric inpatient treatment; (2) currently taking psychiatric medication for symptoms such as hallucinations, delusions, or auditory hallucinations; and (3) being pregnant.

For recruitment, researchers contacted potential participants by telephone, based on their willingness to participate in future research indicated in the consent form for the previous survey [38]. After obtaining written informed consent, 2 trained research team members conducted 1-on-1, face-to-face semistructured interviews, each lasting approximately 2 hours. Interviews were audio recorded and transcribed verbatim using Clova Note (Naver Corp.) [39]. A third researcher verified the quality of the transcriptions. Initially, 21 participants underwent interviews; among them, 10 were selected based on the severity of their traumatic experiences, determined by the number of traumatic events and their PCL-5 scores. We evaluated the average and SD of the participants' demographic characteristics, including age, sex, and duration of residence in South Korea, which reflects the time elapsed since the traumatic event.

### Ethical Considerations

The study protocol was approved by the Institutional Review Board of Yonsei University Health System (approval number Y-2021-0017). Participants were given a gift certificate (equivalent to KRW 50,000 [US $37]) as an incentive and compensation for their participation. We ensured that all study data remained anonymous to protect participants' privacy and confidentiality. Additionally, no images in the manuscript or supplementary materials allow for the identification of individual participants.

### Data Set Labeling

#### Overview

All identifying information, such as the names and residences of the participants, was removed from each interview transcript. Two Korean board–certified mental health professionals—a psychiatrist (BHK) and a clinical psychologist (EK), who were not involved in the data acquisition process—thoroughly reviewed and independently labeled the anonymized transcripts of the 10 participants. They then cross-checked their respective labels on the transcripts and discussed any disagreements. In cases of disagreement regarding the summary labels, we focused on the sections of the summary that were common to both individual labels. We retained only those parts of the summary that both experts agreed were important and removed the rest. When disagreements arose regarding the labeling of symptom sections, we consulted the DSM-5 and the 11th revision of the International Classification of Diseases (ICD-11) to finalize the labels, ensuring they closely matched the definitions of the symptoms. This process yielded 2 types of labels: (1) summary labels and (2) symptom section labels.

#### Summary Label

The summary label comprises a paragraph summarizing the stressors or psychiatric symptoms that likely had a significant impact on each interviewee's life. From each interview, 3 distinct summary labels were generated: an experience summary label, a symptom summary label, and a combined experience and symptom summary label. All summary labels, derived exclusively from the content of the interview transcripts, were presented chronologically, covering the period from childhood to the present.

The word count for both experiences and symptoms is capped at 680 Korean words, allowing for summarized texts of up to 1360 Korean words. This limit corresponds to the maximum token length that the BERT model can process when calculating the BERTScore. For the experience summary labels, the focus was on understanding the interviewee's current psychological state and life history to provide context for the psychiatric symptoms. Priority was given to traumatic and stressful events believed to have influenced psychiatric symptoms. This encompassed a wide range of factors, including childhood personality traits, familial discord, economic and political circumstances, interpersonal relationships in academic and occupational settings, marital status, parental responsibilities, education, religious affiliations, and other life events considered to have particular psychosocial significance.

Symptom summary labels were created to facilitate the identification of psychiatric symptoms and psychological states, thereby aiding in diagnostic decision-making. These labels primarily paraphrased the psychiatric symptoms outlined in the "Symptom Section Label" section, incorporating descriptions of the interviewee's subjective experiences, technical terms from psychopathology and psychology, and terminology consistent with DSM-5 diagnostic criteria. The combined experience and symptom summary labels integrated the 2 preceding summary labels.

#### Symptom Section Label

The symptom section label identifies segments of the interviewee's statements in the transcript that exhibit psychiatric symptoms, along with the corresponding symptom names. The delineation of symptom section labels was limited to segments of the interviewee's utterances that reflected perceptions, cognitions, emotions, and behaviors recognized as psychiatric symptoms that impair daily functioning. The assessment of functional impairment was made within the comprehensive context of the entire transcript.

To enhance the precision of the labeling process and the final labels, the labeling professionals ensured that segments not clinically regarded as psychiatric symptom–related stressors were excluded, even if they could be potentially confusing. Specifically, segments detailing the interviewee's experiences and factual events, discussions of physical injuries or discomfort unrelated to psychiatric symptoms, statements indicating only the duration or recovery of symptoms, accounts of psychiatric symptoms in individuals other than the interviewee, descriptions of general thoughts and emotions typical in cross-cultural adjustment, and reflections on the interviewee's subjective experience of traumatic events were all excluded from the symptom section labels. This process is expected to make the training and evaluation of the aligned LLM more suitable for clinical settings. If an interviewee reiterated the same psychiatric symptom using similar wording, the identical symptom label was applied to encompass all instances within a section.

For example, the statement made by participant P7, "I started to dislike studying, I don't want to study anymore," was recognized as indicating both negative cognitive changes stemming from traumatic experiences and a loss of interest characteristic of depression, resulting in the application of both labels.

The nomenclature of labels adopted a format of symptom abbreviations derived from the symptom lists and definitions of the DSM-5 and ICD-11. In cases where a single symptom encompassed multiple expressions, each manifestation was subcategorized to form distinct labels. For instance, within major depressive disorder, sleep disturbance may manifest as either hypersomnia or insomnia, resulting in the creation of 2 separate labels.

Given that the data set in this study specifically involves North Korean defectors, symptom labels for PTSD from the DSM-5 and complex PTSD from the ICD-11 were developed based on prior research highlighting a propensity for posttraumatic stress symptoms during the resettlement and defection process.

Moreover, symptom labels for depressive disorders, anxiety disorders, and alcohol use disorder—common comorbidities of PTSD identified in the DSM-5—were included. Details of the categories of mental disorders and corresponding symptom labels can be found in Multimedia Appendix 1. For depressive and anxiety disorders, the labels were defined under the assumption that major depressive episodes and panic attacks are representative of their respective disorder categories. Additionally, 1 general anxiety label was defined to encompass clinically significant symptoms associated with anxiety disorders that were not directly traceable to a traumatic experience. As a result, the total number of unique symptoms included in the symptom labels was 36. The final count comprised 515 symptom section labels and 540 symptom-type labels, derived from 10 participant transcripts totaling 375,809 tokens.

### Data Set Split Method

We randomly divided the 10 sets of interview transcripts from North Korean defectors into training, validation, and test data sets. Specifically, 4 sets were allocated to the training data (P4, P11, P14, and P19), 2 sets to the validation data (P5 and P17), and 4 sets to the test data (P3, P7, P9, and P13). For the fine-tuning and in-context learning methods, we utilized the labeled training data, including a total of 184 symptom section labels. For the fine-tuning method, we used the labeled validation data, consisting of 110 symptom section labels.

### Metrics

To delineate sections that indicate evidence of psychiatric symptoms, we assessed the performance of LLMs as follows: For a transcription segment consisting of a single pair of utterances, which includes a ground-truth labeled section, we defined the recall mid-token distance as follows:

$$d := \frac{1}{N}\sum_{i=1}^{N}|a_i - b_i|$$

where $N$, $a_i$, and $b_i$ are defined as below. Let $N$ be the number of ground-truth labeled sections related to psychiatric symptoms within the segment. For the $i$th ground-truth section (eg, the red highlighted parts in Figure 2), where $i = 1, 2, ..., N$, we define $a_i$ as the mid-token index, which represents the index of the token located at the center of the ground-truth section. We define $b_i$ as follows: we compute the mid-token indices of all estimated sections (eg, the yellow highlighted parts in Figure 2) and identify $b_i$ as the computed mid-token index that is closest to $a_i$. According to this definition, we have $d \geq 0$. It is important to note that if no estimated sections are present, we define the recall mid-token distance $d$ as infinity.

The clinical motivation behind the proposed recall mid-token distance is that it is more important to pinpoint the location of the symptom section within a transcript segment than to merely assess the degree of overlap between the ground-truth section and the LLM-estimated section. If the center (mid-token) of the estimated section aligns with that of the ground-truth section, it will aid medical personnel by directing their attention to that specific area, thereby enhancing the clinical consultation process.

For predicting symptom types, we report 4 commonly used metrics in multilabel classification [40]: (1) *accuracy*, (2) *precision*, (3) *recall*, and (4) $F_1$-*score*. For detailed descriptions of these metrics, please refer to Multimedia Appendix 2 (also see [40-42]).

**Figure 2.** An example describing the definition of mid-token distance.

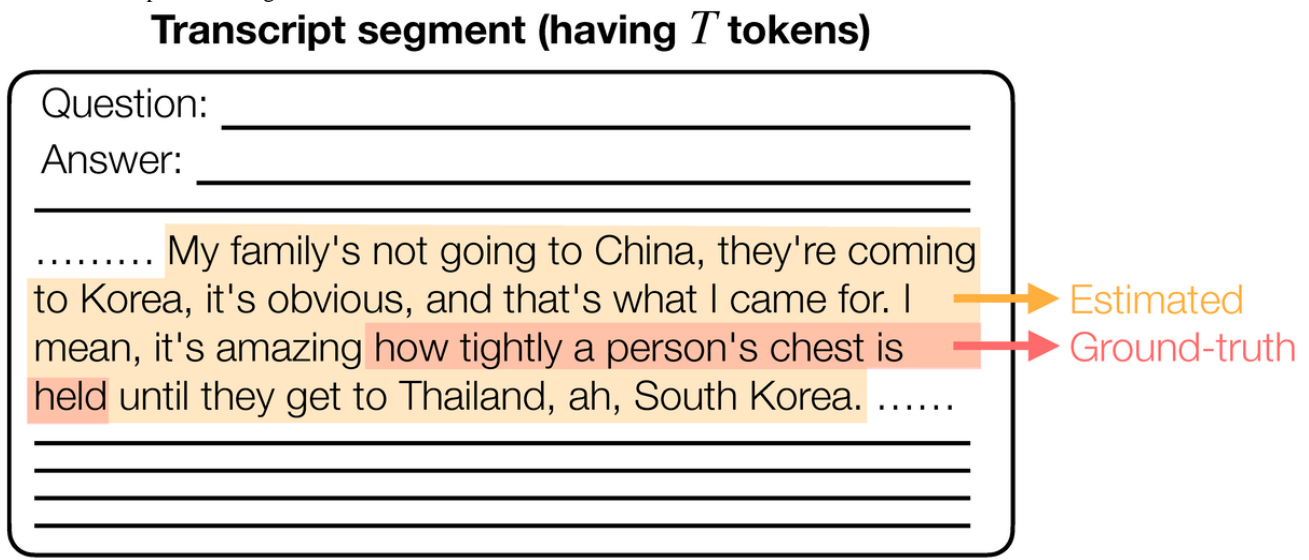

To assess the quality of the summarization of stressors and symptoms by LLMs, we compared the texts generated by LLMs with those produced by human experts. We utilized 2 metrics, BERTScore and G-Eval, which leverage language models to measure the similarity between the 2 texts. BERTScore uses the BERT model, with the derived $F_1$-score ranging from 0 to 1; higher values indicate greater similarity between the texts. For G-Eval, the degree of similarity between 2 texts is directly queried to the GPT-4 model, which provides a similarity score in response. G-Eval evaluates 4 aspects: (1) coherence, (2) consistency, (3) fluency, and (4) relevance, with maximum values of 5, 5, 3, and 5, respectively. The overall G-Eval score is the average of these 4 scores, with a maximum possible score of 4.5.

## Aligning the LLMs

### Task Breakdown and Research Question Alignment

Given the interview transcripts, we aligned the LLMs to perform 3 tasks: (1) extracting stressors from the transcript, (2) delineating symptoms and their indicative sections from the transcript, and (3) summarizing the patients' experiences based on the extracted stressors and symptoms. These tasks address the 2 RQs outlined in the "Introduction" section, where delineating symptoms (RQ1) pertains to the output of the second task, and generating the summary of the interview (RQ2) involves the outputs from all 3 tasks.

### Task 1: Extracting Stressors

For the first task, we extracted patients' stressors or traumatic experiences from the transcript using zero-shot inference with RAG on the GPT-4 Turbo model, as well as zero-shot inference on the GPT-4 Turbo model alone. The stressor extraction module in Figure 3 illustrates the process of extracting stressors from the input transcript. We first divided the input transcript $T$ into $N_{seg}$ disjoint segments ($T_1, T_2, ..., T_{N_{seg}}$), each containing approximately 6000 Korean characters. Subsequently, we used the GPT-4 Turbo model to extract stressors from the contents of each segment $T_i$, yielding the completion response $C_i$, where $i \in \{1, 2, ..., N_{seg}\}$.

**Figure 3.** Two modules for extracting traumatic stressors and symptoms from the transcriptions of interviews using large language models (LLMs).

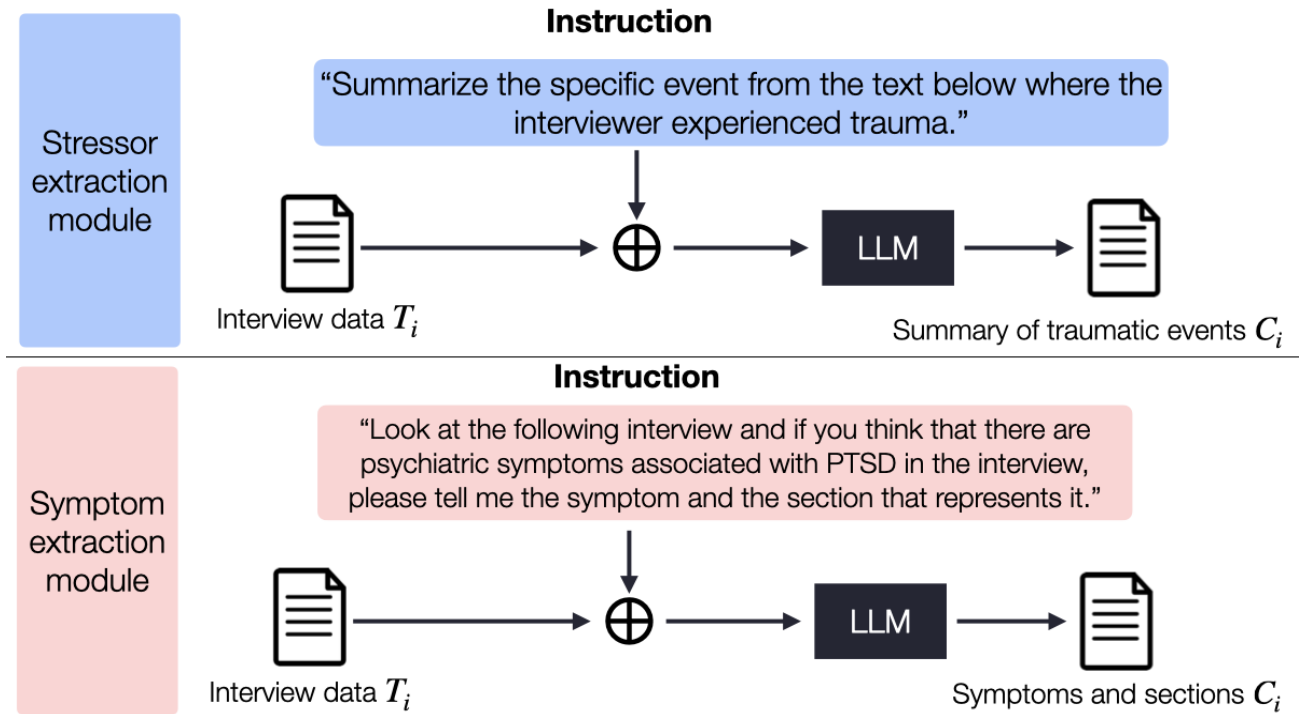

### Task 2: Extracting and Delineating Symptoms

For the second task, we used an LLM to extract and delineate patients' psychiatric symptoms from the provided transcript. This involved inferring (1) which sections of the transcript indicate symptoms and (2) identifying the symptoms themselves. Given the token length limit of the LLM, the transcript is divided into multiple segments, each containing a single exchange between the counselor and the patient. We utilized (1) zero-shot inference, (2) zero-shot inference with RAG, (3) few-shot learning, and (4) fine-tuning to align the LLM with our task and compare their efficacy.

Zero-shot inference involves aligning the LLM with instructional prompts without any parameter updates or explicit in-context examples of the task. The transcription segment and instructions for the LLM to identify psychiatric symptoms are provided as the prompt. In this approach, a list of definitions for all symptoms, which are utilized during the symptom section labeling procedure, is also included in the prompt.

Zero-shot inference with RAG operates similarly to zero-shot inference, but incorporates RAG. In this method, chapters on Trauma and Stressor-Related Disorders from the DSM-5 are used as reference documents, allowing the LLM to retrieve and utilize relevant information from these chapters to formulate a response.

Few-shot learning involves aligning the LLM with instructional prompts and several explicit in-context examples of the task, without updating the model parameters. Specifically, our prompt includes 60 examples of ground-truth (segment, symptom, and section) triplets labeled by mental health professionals. The in-context examples, selected from the training data (P4, P11, P14, and P19), consist of 60 ground-truth (segment, symptom, and section) triplets, prioritizing those with the shortest lengths.

Lastly, fine-tuning involves updating the model parameters of the LLM with a labeled data set to enhance its performance on specific tasks. For fine-tuning, we utilize the ground-truth (segment, symptom, and section) triplet. Specifically, we adjust the LLM's weights to ensure it outputs the symptom and the corresponding section for a given input transcript segment. The validation step is incorporated into the fine-tuning process. To select appropriate hyperparameters, we performed a grid search over the domain of tunable parameters, specifically the learning rate multiplier and the number of epochs, using OpenAI's application programming interface. We evaluated the metrics outlined in the "Delineating Sections and Types of Psychiatric Symptoms" section on the validation data.

Subsequently, we developed the final fine-tuned model using both the training and validation data, applying the best-performing hyperparameter settings, which consisted of 5 epochs and the default learning rate multiplier.

### Task 3: Generating a Summary of the Interview

Finally, we aligned the LLM to generate the interview summary, focusing on the stressors and symptoms obtained from the previous tasks. Three types of summaries were generated: the first version utilized only the extracted stressors from task 1 as input text, while the second version incorporated only the extracted symptoms from task 2. Lastly, both the extracted stressors and symptoms from the previous tasks were combined to create the third version of the summary. Figure 4 illustrates how we generate the different types of summaries using the LLM. Additionally, we performed the same process with RAG to obtain BERTScore and G-Eval scores for each summary.

**Figure 4.** Summarizing patients' stressors and symptoms. LLM: large language model; PTSD: posttraumatic stress disorder.

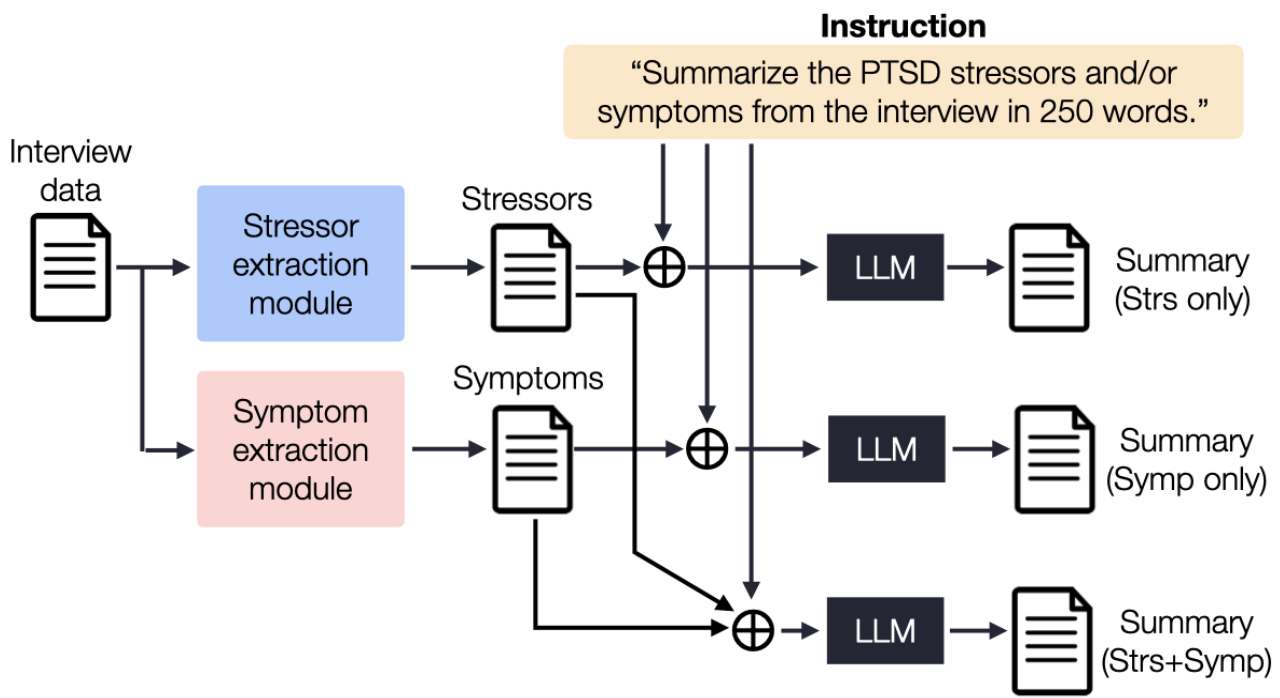

Because of the input token limit of the BERT model, we instructed the LLM to generate concise summaries. We utilized the kcBERT model [43,44], which is specifically trained on Korean texts, to calculate the BERTScore [40]. Two evaluations were conducted for the summaries: BERTScore and G-Eval. In both evaluations, the 3 summary labels outlined in the "Data Set Labeling" section served as reference texts for the corresponding GPT-generated summaries.

For BERTScore, we instructed the GPT-4 Turbo model to shorten the summary labels for stressors and symptoms, as the BERT model has an input token limit. We obtained the BERTScore ($F_1$-score) as a quantitative evaluation metric to assess the similarity between the summaries generated by human experts and those produced by the GPT model.

For the G-Eval evaluation, we obtained scores for (1) coherence, (2) consistency, (3) fluency, and (4) relevance. These scores served as quantitative metrics to assess the quality of the GPT-generated summaries and their similarity to the summary labels. Note that the evaluation was conducted using the gpt-4-0314 model, as we found that G-Eval does not yield consistent results when switching models. Therefore, we utilized the gpt-4-0314 model, which is referenced as the GPT-4 model in the paper [36].

### Retrieval-Augmented Generation

RAG is a method that enhances LLMs by integrating data from external knowledge sources, thereby improving the accuracy and contextual relevance of their responses. This technique enables LLMs to access up-to-date and domain-specific information, resulting in more reliable and pertinent answers without the need for retraining the model. RAG is known to be particularly beneficial for enhancing the factuality of LLMs [45], especially in cases where the generated output necessitates specific domain knowledge. In this study, we embedded the chapters on Trauma and Stressor-Related Disorders from the DSM-5 [46] as reference documents for the LLM to retrieve and utilize during the generation process. RAG was used for 2 primary tasks: extracting stressors and delineating symptoms. To manage long texts, we specifically used the *RecursiveCharacterTextSplitter* function in Langchain [47], followed by embedding the split texts using the text-embedding-ada model developed by OpenAI [48]. Afterward, we used Facebook AI Similarity Search (FAISS) [49] to index and retrieve the embeddings relevant to the given query.

### Understanding the Performances of Distinct Methods on Different Tasks

We conducted additional experiments to gain insights into why certain methods outperform others in each task. Specifically, we focused on the symptom estimation task and the section estimation task (results presented in Multimedia Appendix 3). According to the table and figures, the "fine-tuned GPT-3.5 Turbo model" outperforms the "zero-shot GPT-4 Turbo model" in the *symptom* estimation task, while the latter surpasses the former in the *section* estimation task. To better understand this discrepancy, we conducted additional experiments as outlined below.

The basic idea of our new experiment is to measure performance on 2 different sets of segments: the first option evaluates all segments in the transcript, while the second option focuses solely on the "positive segments," which are defined as those containing at least one ground-truth label. For these 2 setups, we compare 3 models (fine-tuned GPT-3.5, zero-shot GPT-4, and in-context learned GPT-4) across 2 tasks (symptom estimation and section estimation) in Multimedia Appendix 4, which helps to elucidate the discrepancy.

## Results

### Addressing Research Questions: LLM Performance in Symptom Delineation and Summarization

In this section, we present results that address our RQs (RQ1 and RQ2) stated in the "Introduction" section. RQ1 is addressed in the "Delineating Sections and Types of Psychiatric Symptoms" section, where we demonstrate the performance of LLMs in (1) identifying the sections of the conversation that indicate psychiatric symptoms and (2) predicting the corresponding symptoms. RQ2 is discussed in the "Summarizing Stressors and Symptoms From the Interview" section, where we evaluate how effectively LLMs summarize patients' stressors and symptoms derived from the interviews. In particular, we compare the summaries generated by LLMs with those written by human experts. It is important to note that the transcript data used in our experiments are in Korean, meaning both the inputs and the outputs are in Korean. In the manuscript, we provide the English version, which has been translated using DeepL [50]. For reproducibility, we share our code in a public GitHub repository [51]. Details of the prompts used in our experiments are provided in Multimedia Appendix 5.

### Demographic Characteristics of Patients

Our study utilizes transcripts from 10 patients, including 9 females and 1 male, with an average age of 44.4 years (SD 8.4) years and an average duration of residence in South Korea of 12.4 (SD 4.5) years. Detailed demographic characteristics of the participants are presented in Table 1.

Table 1. Demographic characteristics of the patients.

| ID (P#) | Gender | Age | Initial attempt to defect from North Korea (year) | Residency years in South Korea | Number of traumatic events | PCL-5[a] score |
|---|---|---|---|---|---|---|
| P3 | Female | 37 | 2004 | 13 | 7 | 54 |
| P4 | Female | 37 | 1998 | 12 | 11 | 64 |
| P5 | Female | 59 | 2000 | 19 | 8 | 74 |
| P7 | Male | 53 | 2011 | 11 | 12 | 41 |
| P9 | Female | 60 | 1999 | 17 | 11 | 62 |
| P11 | Female | 51 | 1998 | 13 | 13 | 39 |
| P13 | Female | 46 | 2008 | 12 | 12 | 65 |
| P14 | Female | 47 | 2003 | 16 | 9 | 68 |
| P17 | Female | 37 | 2001 | 14 | 15 | 61 |
| P19 | Female | 33 | 2006 | 7 | 9 | 40 |

[a]PCL-5: DSM-5's Posttraumatic Stress Disorder Checklist.

### Delineating Sections and Types of Psychiatric Symptoms

In this section, we present the performance of LLMs in estimating (1) the transcript sections related to psychiatric symptoms and (2) the names of the corresponding symptoms. The results are reported for 3 different methods of utilizing LLMs: zero-shot inference, zero-shot inference with RAG, and few-shot learning (eg, in-context learning) and fine-tuning.

We evaluate the performance of LLMs on transcripts from 4 patients (denoted as P3, P7, P9, and P13), which include a total of 246 symptom section labels. Each transcript consists of multiple pairs of utterances between the interviewer and the interviewee. For each pair of utterances, we first instruct the LLM to determine whether the pair contains any content indicative of psychiatric symptoms, and then to estimate the symptoms.

Table 2 presents the recall mid-token distance $d$ of the sections estimated by the GPT-4 Turbo model in the zero-shot inference setting. We assess the performance of LLMs using transcripts from 4 patients (denoted as P3, P7, P9, and P13). The recall mid-token distance $d$ is computed for 102 transcript segments that contain labeled symptom sections, categorizing the segments based on the range of $d$ in Table 2. Notably, among the 102 segments, 73 segments have a distance $d<20$ when evaluated using the zero-shot inference on the GPT-4 Turbo model.

**Table 2.** Performance of the zero-shot inference setting using the GPT-4 Turbo model on delineating the ground-truth labels, as measured by the recall mid-token distance ($d$).

| Range | Frequency |
| --- | --- |
| $0 \leq d < 10$ | 58 |
| $10 \leq d < 20$ | 15 |
| $20 \leq d < 50$ | 13 |
| $d \geq 50$ | 16 |

Table 3 presents examples of the ground-truth sections alongside the estimated sections, as well as the corresponding recall mid-token distance $d$. It is important to note that when $d=0$, the estimated section is identical to the ground-truth section, indicating accurate prediction. By contrast, for examples with larger values of $d$, the overlap between the 2 sections diminishes. Specifically, in the segment where $d=0$, the model accurately predicts the section. In cases with a value of $d=6$, the predicted section encompasses the ground-truth section. Additionally, we provide a histogram of the mid-token distances measured across different methods in Multimedia Appendix 3.

**Table 3.** Examples of comparison of (1) labeled (ground-truth) sections related to symptoms and (2) sections estimated by large language models within given transcript segments. The estimation becomes more accurate (ie, there is a greater overlap between the ground-truth and estimated sections) as the corresponding mid-token distance ($d$) decreases.

| Recall mid-token distance $d$ | Ground-truth section | Estimated section |
| --- | --- | --- |
| 0 | *But when I dream about it, I dream about the scene of my escape, the scene of my escape from North Korea, the scene of my escape from the police, and I still dream about it.* | *But when I dream about it, I dream about the scene of my escape, the scene of my escape from North Korea, the scene of my escape from the police, and I still dream about it.* |
| 2 | *Memory I don't really want to think about* | *I don't know, I haven't pulled it out in a long time, and it's actually a memory I don't really want to think about. Yeah.* |
| 6 | *Yes. That's hard and scary too.* | *It sounds like it's hard for you to be intimate with guys and have new relationships and stuff like that. P3: Yeah. That's hard and scary too.* |
| 31 | *That's what I still think about now, why did I say that, when he's gone, why did I say that, and that's what I regret.* | *Never the things of my heart. I am unjust. My heart is broken. I'm hurting. I'm just not expressing it.* |
| Infinity | *It's because we're conditioned to think that anyone in black is someone who's out to get us.* | None |

Table 4 illustrates the performance of LLMs in estimating the symptoms of patients. We assessed the performance of LLMs on transcripts from 4 patients (denoted by P3, P7, P9, and P13) and report 4 commonly used metrics for multilabel classification [40]: (1) accuracy, (2) precision, (3) recall, and (4) F1-score. Additionally, to evaluate the performance of LLMs in estimating negative segments, we report the negative predictive values. The results confirm that both fine-tuning (which utilizes training data) and RAG (which incorporates external documents) provide a performance advantage over the zero-shot inference setting in the GPT-4 Turbo model.

Table 4. Performance of large language models on estimating symptoms based on the interview data.[a]

| Model | Method | Accuracy | Precision (positive predictive value[b]) | Negative predictive value[b] | Recall | $F_1$-score |
| --- | --- | --- | --- | --- | --- | --- |
| GPT-3.5 Turbo | Fine-tuning | 0.817 (0.002) | 0.828 (0.002) | 0.866 (0.001) | 0.818 (0.001) | 0.821 (0.002) |
| GPT-4 Turbo | In-context learning | 0.537 (0.008) | 0.551 (0.009) | 0.989 (0.003) | 0.550 (0.007) | 0.546 (0.008) |
| GPT-4 Turbo | Zero-shot | 0.644 (0.004) | 0.649 (0.003) | 0.965 (0.011) | 0.681 (0.002) | 0.657 (0.003) |
| GPT-4 Turbo | Zero-shot (with RAG[c]) | 0.708 (0.005) | 0.715 (0.007) | 0.954 (0.000) | 0.745 (0.005) | 0.722 (0.005) |

[a]For each setting, we report the mean and the SD of score values for 3 trials. As we use nonzero temperature parameters of large language models, the performance varies among different trials.
[b]The definitions of negative predictive value and positive predictive value are given in Multimedia Appendix 2.
[c]RAG: retrieval-augmented generation.

Table 5 presents examples of symptoms estimated by the fine-tuned GPT-3.5 Turbo model for each transcript segment. The table highlights error types, including instances where symptoms are present but predicted to be absent, as well as cases where only some symptoms are predicted despite multiple symptoms being present.

Table 5. Comparison between the ground-truth symptoms (labeled by a human expert) and the symptoms estimated by the fine-tuned GPT-3.5 Turbo model, with 4 metrics for each transcript segment.

| Transcript segment | Ground-truth symptom | Estimated symptom | Accuracy, precision, recall, and $F_1$-score |
| --- | --- | --- | --- |
| ...Yes, there is such a stereotype. But in reality, as I walk around so energetically, people start imitating the way I walk, saying things like "You're like a gangster,"...Anyway, being swept up in that group, within the circle of physical education, I think I just showed my true personality. | None | None | 1, 1, 1, and 1 |
| ...Back then, I felt so trapped and thought that maybe I shouldn't have come from North Korea. Such thoughts crossed my mind....In reality, I couldn't live in North Korea anymore. It was really tough back then, especially while I was in China. | None | Negative change in mood | 0, 0, 0, and 0 |
| ...Yes, so when I first came to South Korea, the sound of ambulances was so overwhelming. Every time I heard an ambulance, I would instinctively jump and move to hide my body. In the past, I would unconsciously find a place to hide whenever I heard an ambulance siren. | Arousal | Arousal | 1, 1, 1, and 1 |
| ...Instead, when I go home in the evening, I can't sleep. If I spend the day feeling a certain way, it keeps me up at night. So, I calm myself with a drink. After having a drink, I'm able to sleep a bit.... | Alcohol dependence and insomnia | Insomnia | 0.5, 1, 0.5, and 0.67 |
| ...But it feels like a vicious cycle. Those experiences from childhood, marriage, childbirth, and then the challenges in communication and culture – it all stems from experiences I had when I was young....I made choices irresponsibly, without loving myself, just thinking I need to be protected, and just making choices haphazardly. | Negative self-image and negative change in cognition | None | 0, 0, 0, and 0 |

## Summarizing Stressors and Symptoms From the Interview

Table 6 presents the quantitative performance of the GPT-4 Turbo model in generating summaries for 4 patients (test data: P3, P7, P9, and P13). The results are derived from zero-shot inference using the GPT-4 Turbo model to extract stressors (*Strs*) and symptoms (*Symp*) from the input transcript. We compared 3 versions of the summaries: those containing only stressors, only symptoms, and both stressors and symptoms. The overall score is the average of 4 individual scores, with 4.5 being the maximum. As a G-Eval score above 3.8 is considered human-level [36], Table 6 demonstrates that the quality of the summaries generated by the LLM is reasonably high. It can be observed that the summaries are of the highest quality when *both* stressors and symptoms are included, compared with when only stressors or *only* symptoms are used.

**Table 6.** Evaluation of summaries generated by the GPT-4 Turbo model for patients.

| Summaries | G-Eval[a], mean (SD) | | | | | BERT[b,c], mean (SD) |
| --- | --- | --- | --- | --- | --- | --- |
| | Coherence | Consistency | Fluency | Relevance | Overall | Score |
| Strs[d] | 4.22 (0.19) | 4.02 (0.33) | 1.55 (0.60) | 4.21 (0.38) | 3.50 (0.26) | 0.51 (0.03) |
| Symp[e] | 4.43 (0.21) | 4.34 (0.71) | 1.15 (0.12) | 4.42 (0.38) | 3.59 (0.28) | 0.54 (0.02) |
| Strs+Symp[f] | 4.66 (0.08) | 4.73 (0.07) | 2.16 (0.71) | 4.67 (0.13) | 4.01 (0.17) | 0.58 (0.01) |
| Strs (with RAG[g]) | 4.31 (0.28) | 3.75 (0.85) | 1.45 (0.36) | 4.30 (0.28) | 3.41 (0.28) | 0.49 (0.02) |
| Symp (with RAG) | 4.09 (0.41) | 3.92 (0.87) | 1.53 (0.69) | 4.09 (0.57) | 3.40 (0.48) | 0.52 (0.03) |
| Strs+Symp (with RAG) | 4.51 (0.08) | 4.69 (0.09) | 2.11 (0.49) | 4.51 (0.17) | 3.96 (0.17) | 0.58 (0.02) |

[a]G-Eval measures the coherence, consistency, fluency, and relevance of GPT's summary.
[b]BERT: Bidirectional Encoder Representations from Transformers.
[c]BERTScore measures the similarity between the summaries generated by GPT-4 Turbo and those from a human expert.
[d]Strs uses estimated stressors only.
[e]Symp uses estimated symptoms only.
[f]Strs+Symp uses both estimated stressors and symptoms.
[g]RAG: retrieval-augmented generation.

We also tested the effect of using RAG on summarization performance. As shown in Table 6, RAG did not lead to a significant increase in G-Eval scores.

For qualitative assessment, Multimedia Appendix 6 presents the summary texts generated for patient P9 by a human expert and the LLMs. We compared 3 versions: the summary created by the human expert, the GPT-4 Turbo model, and the GPT-4 Turbo model with RAG.

### Comparison of Performances Measured for Different Sets of Transcript Segments

In Multimedia Appendix 4, "positive segments" refer to transcript segments that contain at least one ground-truth label. For the section estimation task, we reported the average recall mid-token distance $d$, which is defined only for positive segments. As the recall mid-token distance is not measured for negative segments (ie, transcript segments without a ground-truth label), we cannot calculate results for "all segments," which include both positive and negative segments. As a result, some cells in Multimedia Appendix 4 are marked with a "—."

Note that even in positive segments containing at least one ground-truth label, there are instances where the LLMs fail to identify the label and do not output any corresponding estimated section. We excluded such cases where $d=\infty$ and reported the proportion of these segments in the "Ratio of the absence of the estimated sections" column. The average $d$ was calculated for the remaining segments.

As shown in Table 4 and Multimedia Appendix 3, no single method consistently outperforms the others across both the symptom estimation task and the section estimation task. To better understand the discrepancy in performance between these tasks, Multimedia Appendix 4 presents the results measured on 2 different sets of segments: (1) all segments in the transcript and (2) positive segments, which contain at least one ground-truth label.

By including the measurement on "positive segments," Multimedia Appendix 4 conveys a consistent message:

- The fine-tuned GPT-3.5 Turbo model performs best when evaluated on "all segments" but performs the worst when measured on "positive segments" only.
- The zero-shot GPT-4 Turbo model, by contrast, performs best when evaluated on "positive segments" only, but performs the worst when measured on "all segments."

By investigating the outcomes of LLMs, we observed that the fine-tuning method excels at identifying the absence of ground-truth labels in the conversation. However, it performs poorly in classifying the symptom type and the corresponding section when the transcript segment contains at least one ground-truth label.

## Discussion

### Summary of Main Findings

In this paper, we investigated the alignment of LLMs to support clinical practice in psychiatric evaluations and validated their performance using interview transcript data. Specifically, we aligned the LLMs to generate reports on (1) delineating sections and types of psychiatric symptoms in patients by using zero-shot and few-shot learning, along with RAG and fine-tuning; and (2) summarizing the stressors, symptoms, or both from the interviews. The results align with recent evidence suggesting that LLMs can perform remarkably well on structured medical question-answering benchmarks [21,28,33]. They support the promising potential of LLMs as practical aids in the clinical field [24-27,29,52], particularly in psychiatry, as demonstrated in this study.

### Interpretations and Implications of Main Findings

In the psychiatric assessment and interview process, certain utterances are particularly crucial as they indicate the patient's symptoms and signs. Distinguishing whether these utterances correspond to significant symptoms and signs provides clinicians with insights into areas that require closer examination during psychiatric interviews. This can assist in clinical practice by providing clinicians with a second opinion on which parts of the interview to review and by enhancing interpretability and reliability through explanations of why certain symptoms are suggested to be present by the language model (LLM). Accordingly, we validated the LLMs' ability to identify dialogue segments indicative of specific psychopathologies and to suggest the corresponding psychopathological conditions.

When delineating symptoms, we introduced the "recall mid-token distance" as a quantitative metric for evaluating prediction quality. We posited that, in a real clinical practice setting, it is crucial to indicate where the clinician should focus on rather than to make a precise prediction of symptom segments with the LLM. Thus, the recall mid-token distance is designed to measure how close the center of the LLM-suggested segment is to the ground-truth segment labeled by professionals. Given that the zero-shot–prompted GPT-4 Turbo model was able to delineate 70% of the tested segments, we can conclude that it is reasonably effective at suggesting the symptom segments on which clinicians should focus.

The LLM also demonstrated a high level of accuracy in suggesting which symptom or psychopathology the predicted segment relates to. Specifically, the fine-tuned GPT-3.5 Turbo model achieved an accuracy of 0.817 in the multiclass classification of symptom labels. This high accuracy indicates that the LLM can effectively highlight which symptoms should be considered from the patient's utterances for psychiatrists. Although the final decision rests with the clinicians, such suggestions are intended to support the decision-making process by providing an auxiliary opinion.

## Limitations

We acknowledge 4 major limitations of our work. First, our experiments were conducted using an in-house data set that is limited to a specific group of patients. This limitation may restrict the generalizability of our results to other psychiatric disorders due to a lack of external validity. To address this limitation, we created simulated samples and conducted additional experiments. The process of creating the simulated samples and the experimental results are detailed in Multimedia Appendix 7 (see also [53]). As another potential solution, we suggest considering the use of publicly available therapy dialogue data sets for future studies. However, using a private data set ensures that the data were not utilized during the training of proprietary LLMs such as GPT. Second, we did not evaluate our methods on real-time interviews; instead, we based our evaluation on transcripts derived from audio recordings. Implementing a pipeline that leverages speech recognition technology for use in real-world clinical situations is a promising avenue for future work. Third, while the recall mid-token distance metric was proposed from a clinical perspective, a deeper understanding of its effectiveness is necessary. Lastly, the performance of our method is contingent on how we labeled the transcripts used as training data. As our current results are based on our own labeling method for segments related to psychiatric symptom–related stressors, a more thorough analysis of the impact of labeling on the performance of LLMs should be conducted.

## Conclusions

We proposed a novel pipeline for delineating sections and types of psychiatric symptoms, as well as summarizing symptoms and traumatic experiences from patients' utterances. We anticipate that this automated extraction and summarization can enhance the clinical workflow for psychiatrists. For example, the generated summaries can assist psychiatrists by providing quick recalls of significant patient mentions or serve as drafts for clinical notes, thereby saving time. This approach is particularly valuable in low-resource settings or during traumatic emergencies, such as natural disasters, wars, and acts of terror, where the demand for mental health services often exceeds available resources. In these scenarios, LLMs could provide essential preclinical information that aids mental health specialists in making diagnostic and treatment decisions. However, there are concerns that relying heavily on this method could complicate in-depth counseling sessions that require building rapport through face-to-face interactions, potentially hindering the development of deeper therapeutic relationships. Such relationships are crucial and are often facilitated by direct personal engagement. Therefore, while the initial use of LLMs should be to support mental health specialists, it is essential to balance technological assistance with the need for personal interaction in therapeutic settings.

## Acknowledgments

This research was supported by the MSIT (Ministry of Science and ICT), Korea, under the ICAN (ICT Challenge and Advanced Network of HRD) support program (grant RS-2023-00259934), supervised by the IITP (Institute for Information & Communications Technology Planning & Evaluation); the Basic Science Research Program through the National Research Foundation of Korea (NRF), funded by the Ministry of Education (grants NRF-2022R1I1A1A01069589 and 2019R1I1A2A01058746); the NRF grant funded by the Korean government (MSIT; grant RS-2024-00341793); and a National Research Foundation of Korea (NRF) grant funded by the Korea government (MSIT; grant RS-2024-00345351). We express our special thanks to Ocksim Kim and Minjeong Ko for conducting interviews with North Korean defectors, following rigorous training on qualitative interviews. We also acknowledge Dr. Seong Gyu Choi for reviewing the quality of the transcriptions.

## Data Availability

Because of the sensitive nature of this study involving extremely vulnerable North Korean defectors, and in strict adherence to ethical guidelines, the deidentified data will not be available for public sharing. However, Python codes used for data analyses are available in the GitHub Repository [51].

## Authors' Contributions

JYC, J Sohn, BHK, and SHC contributed to the study's concept and design. J So, JC, J Sohn, BHK, and SHC drafted the manuscript. All authors made critical revisions to the manuscript, contributing important intellectual content. JYC and SHC contributed to data set acquisition. EK and BHK were responsible for data labeling. J So, JC, and JN wrote the Python code for running experiments on large language models. SHC obtained funding for this study. J Sohn, BHK, and SHC are co-corresponding authors and supervised the entire study. J So and JC contributed equally in running experiments and writing the draft of the paper. All authors accept final responsibility for submitting the manuscript for publication.

## Conflicts of Interest

None declared.

## Multimedia Appendix 1

Categories of mental disorders and the corresponding symptom labels.
[[PDF File (Adobe PDF File), 46 KB](#)-[Multimedia Appendix 1](#)]

## Multimedia Appendix 2

Details of the experimental settings.
[[PDF File (Adobe PDF File), 127 KB](#)-[Multimedia Appendix 2](#)]

## Multimedia Appendix 3

Histograms of the recall mid-token distances measured for large language models (LLMs).
[[PDF File (Adobe PDF File), 2062 KB](#)-[Multimedia Appendix 3](#)]

## Multimedia Appendix 4

Performance of 3 different methods on symptom/section estimation tasks, evaluated on (1) all segments and (2) positive segments. We conducted experiments over 3 trials and report the mean and SD of the results.
[[DOCX File , 14 KB](#)-[Multimedia Appendix 4](#)]

## Multimedia Appendix 5

Prompts used in experiments.
[[PDF File (Adobe PDF File), 75 KB](#)-[Multimedia Appendix 5](#)]

## Multimedia Appendix 6

Comparison of the summaries generated by human experts, the GPT-4 Turbo model, and the GPT-4 Turbo model using retrieval-augmented generation.
[[PDF File (Adobe PDF File), 55 KB](#)-[Multimedia Appendix 6](#)]

## Multimedia Appendix 7

Additional experiments on simulated data generated by large language models.
[[PDF File (Adobe PDF File), 117 KB](#)-[Multimedia Appendix 7](#)]

## Abbreviations

**AI:** artificial intelligence
**BERT:** Bidirectional Encoder Representations from Transformers

**BERTScore:** Bidirectional Encoder Representations from Transformers Score
**DSM-5:** Diagnostic and Statistical Manual of Mental Disorders, Fifth Edition
**FAISS:** Facebook AI Similarity Search
**G-Eval:** Generative Evaluation
**ICD-11:** the eleventh revision of the International Classification of Diseases
**LLM:** large language model
**Med-PaLM:** Medical-Pathways Language Model
**PCL-5:** DSM-5's Posttraumatic Stress Disorder Checklist
**PTSD:** posttraumatic stress disorder
**RAG:** retrieval-augmented generation
**RQ:** research question